\documentclass[conference]{IEEEtran}
\pdfoutput=1

\usepackage{times}

\usepackage[numbers,sort&compress]{natbib}
\usepackage{multicol}
\usepackage[bookmarks=true]{hyperref}
\hypersetup{colorlinks=true,linkcolor=blue}

\usepackage{graphicx}
\usepackage{amssymb}
\usepackage{amsmath}
\usepackage{color}
\usepackage{bbm}
\usepackage{booktabs}
\usepackage{caption}
\usepackage{dsfont}
\usepackage{verbatim}
\usepackage{pgfplots}

\usepackage{algorithmicx}
\usepackage[noend]{algpseudocode}
\usepackage{algorithm}
\usepackage[T1]{fontenc}
\newcommand{\figref}[1]{Figure~\ref{#1}}
\newcommand{\secref}[1]{Section~\ref{#1}}
\newcommand{\tabref}[1]{Table~\ref{#1}}

\algnewcommand\algorithmicdefinitions{\textbf{Definitions:}}
\algnewcommand\Definitions{\item[\algorithmicdefinitions]}
\renewcommand{\algorithmiccomment}[1]{{\color{gray}\raisebox{1px}{\texttt{\guillemotright}} #1}}
\algnewcommand{\LineComment}[1]{\Statex \hskip\ALG@thistlm \algorithmiccomment{#1}}
\algrenewcommand\alglinenumber[1]{\footnotesize #1:}
\algrenewcommand\algorithmicindent{1.0em}%
\makeatletter
\newcommand{\StatexIndent}[1][3]{%
  \setlength\@tempdima{\algorithmicindent}%
  \Statex\hskip\dimexpr#1\@tempdima\relax}

\usepackage{etoolbox}
\makeatletter
\patchcmd{\@makecaption}
  {\scshape}
  {}
  {}
  {}
\makeatother

\newcommand{\eat}[1]{\ignorespaces}
\newcommand{\reals}{\mathds{R}}

\newcommand{\allinstructions}{\mathcal{X}}

\newcommand{\state}{s}
\newcommand{\allstates}{\stateset}

\newcommand{\position}{p}
\newcommand{\posseq}{\Xi}
\newcommand{\lmpos}{\position_C}
\newcommand{\goalpos}{\position_g}
\newcommand{\successregion}{r_g}
\newcommand{\stopthres}{D_{stop}}
\newcommand{\hyperstopthres}{6\text{ meters}}

\newcommand{\context}{\tilde{\state}}

\newcommand{\act}[1]{{\tt #1}}
\newcommand{\action}{a}
\newcommand{\expertaction}{a^*}
\newcommand{\allactions}{\actionset}
\newcommand{\stopaction}{\act{STOP}}

\newcommand{\imagefunc}{\textsc{Img}}
\newcommand{\posefunc}{\textsc{Loc}}

\newcommand{\vanilladagger}{\textsc{DAgger}}
\newcommand{\daggerfm}{\textsc{DAggerFM}}

\newcommand{\policy}{\pi}
\newcommand{\oracle}{\pi^*}

\newcommand{\stateset}{\mathcal{S}}

\newcommand{\actionset}{\mathcal{A}}

\newcommand{\instruction}{u}

\newcommand{\policystatedist}{d_\pi}
\newcommand{\policysampledist}{d_{\hat{\pi}}}

\newcommand{\actiondistance}{D_a}

\newcommand{\jaction}{J_{{\rm act}}}
\newcommand{\jlang}{J_{{\rm lang}}}
\newcommand{\jclass}{J_{{\rm percept}}}
\newcommand{\jrel}{J_{{\rm ground}}}
\newcommand{\jgoal}{J_{{\rm plan}}}

\newcommand{\weightclass}{\lambda_v}
\newcommand{\weightlang}{\lambda_l}
\newcommand{\weightrel}{\lambda_g}
\newcommand{\weightgoal}{\lambda_p}

\newcommand{\velfwd}{v}
\newcommand{\velang}{\omega}
\newcommand{\stopprob}{p_{{\rm stop}}}
\newcommand{\velfwdstar}{v^*}
\newcommand{\velangstar}{\omega^*}
\newcommand{\stopprobstar}{p_{{\rm stop}}^*}

\newcommand{\image}{I}
\newcommand{\pose}{P}
\newcommand{\pos}{p}
\newcommand{\rot}{\theta}

\newcommand{\object}{O}

\newcommand{\worldframe}{W}
\newcommand{\camframe}{C}
\newcommand{\robotframe}{R}

\newcommand{\fmcam}{F^\camframe}

\newcommand{\fmworld}{F^\worldframe}
\newcommand{\maskworld}{M^\worldframe}
\newcommand{\smworld}{S^\worldframe}

\newcommand{\rmworld}{R^\worldframe}
\newcommand{\rmcam}{R^\camframe}
\newcommand{\rmrobot}{R^\robotframe}
\newcommand{\gmworld}{G^\worldframe}
\newcommand{\gmcam}{G^\camframe}
\newcommand{\gmrobot}{G^\robotframe}

\newcommand{\raycast}{\textsc{Raycast}}
\newcommand{\resblock}{\textsc{ResBlock}}

\newcommand{\embed}{\phi}
\newcommand{\param}{\Theta}
\newcommand{\weights}{W}
\newcommand{\bias}{b}

\newcommand{\resnet}{\textsc{ResNet} }
\newcommand{\lstm}{\textsc{LSTM} }

\newcommand{\adam}{\textsc{ADAM}}

\newcommand{\lab}{{\rm label}}
\newcommand{\spat}{{\rm spatial}}

\newcommand{\convlab}{\textsc{Conv}}
\newcommand{\convspat}{\textsc{Conv}}
\newcommand{\mlp}{\textsc{DenseMLP}}
\newcommand{\leakyrelu}{\textsc{LeakyReLU}}

\newcommand{\modelname}{\textsc{GSMN}}
\newcommand{\fullmodelname}{\text{Grounded Semantic Mapping Network}}
\newcommand{\modelnogoal}{\textsc{GSMN w/o $\jgoal$}}

\newcommand{\blimg}{\textsc{GS-FPV}}
\newcommand{\blimgrec}{\textsc{GS-FPV-MEM}}

\newcommand{\best}{\textbf}

\newcommand{\supdataset}{\mathcal{D}^*}

\newcommand{\numlandmarks}{|\object|}
\newcommand{\leftside}{\textit{left}}
\newcommand{\rightside}{\textit{right}}
\newcommand{\frontside}{\textit{front}}
\newcommand{\backside}{\textit{back}}

\newcommand{\avgnumsteps}{\text{30}}

\newcommand{\trainsetsize}{\text{3500}}
\newcommand{\devsetsize}{\text{750}}
\newcommand{\testsetsize}{\text{750}}
\newcommand{\prunedtestsetsize}{\text{690}}

\newcommand{\lambdaaux}{\text{0.1}}

\newcommand{\maxvelx}{\text{1.6 m/s}}
\newcommand{\maxvelw}{\text{2.44 rad/s}}

\begin{document}

\title{Following High-level Navigation Instructions on a Simulated Quadcopter with Imitation Learning}

\author{\authorblockN{Valts Blukis\authorrefmark{1}\authorrefmark{2}, Nataly Brukhim\authorrefmark{3}\footnote{Work done at Cornell Tech}, Andrew Bennett\authorrefmark{1}\authorrefmark{2}, Ross A. Knepper\authorrefmark{1}, Yoav Artzi\authorrefmark{1}\authorrefmark{2}}
\authorblockA{\authorrefmark{1}Department of Computer Science, Cornell University, Ithaca, New York, USA}
\authorblockA{\authorrefmark{2}Cornell Tech, Cornell University, New York, New York, USA}
\authorblockA{\authorrefmark{3}Tel Aviv University, Tel Aviv-Yafo, Israel}
\authorblockA{Email: \{valts, awbennett, rak, yoav\}@cs.cornell.edu, natalybr@mail.tau.ac.il}
}

\maketitle

\begin{abstract}
We introduce a method for following high-level navigation instructions by mapping directly from images, instructions and pose estimates to continuous low-level velocity commands for real-time control.
The Grounded Semantic Mapping Network ($\modelname$) is a fully-differentiable neural network architecture that builds an explicit semantic map in the world reference frame by incorporating a pinhole camera projection model within the network. The information stored in the map is learned from experience, while the local-to-world transformation is computed explicitly.
We train the model using $\daggerfm$, a modified variant of $\vanilladagger$ that trades tabular convergence guarantees for improved training speed and memory use.
We test $\modelname$ in  virtual environments on a realistic quadcopter simulator and show that incorporating an explicit mapping and grounding modules allows $\modelname$ to outperform strong neural baselines and almost reach an expert policy performance. Finally, we analyze the learned map representations and show that using an explicit map leads to an interpretable instruction-following model.
\end{abstract}

\IEEEpeerreviewmaketitle

\section{Introduction}
\label{sec:intro}

\begin{figure}
\centering
\includegraphics[scale=0.124]{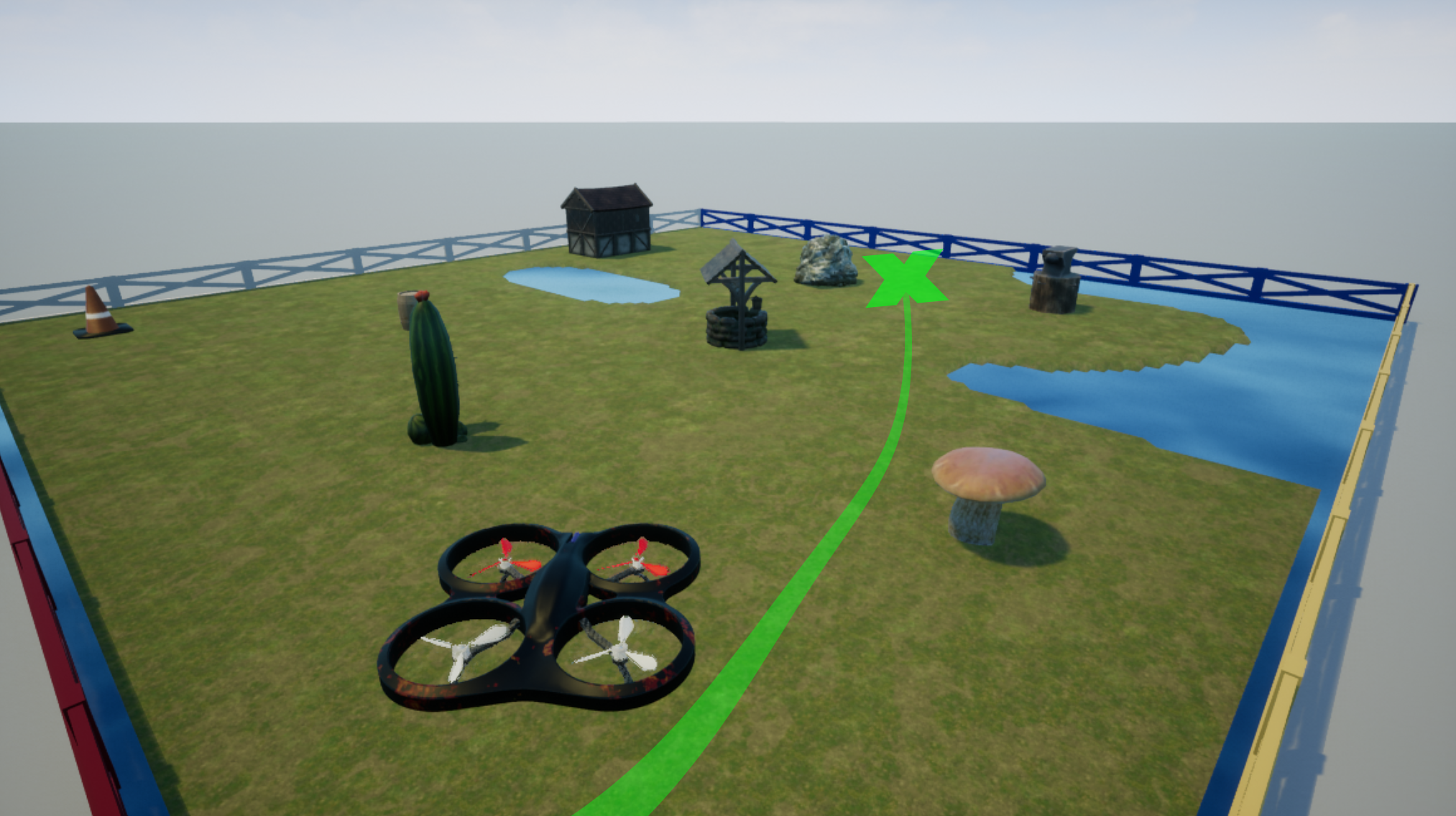}
\fbox{
\textit{Go to the right side of the rock}}
\caption{High-level instruction in our navigation environment, and an illustration of the goal position and  trajectory that the agent must infer and follow given its observations.}
\label{fig:main}
\end{figure}

Autonomous navigation from high-level instructions requires solving perception, planning and control challenges. 
Consider the navigation task in \figref{fig:main}. 
To complete the task, a quadcopter must reason about the instruction, observations of the environment, and the sequence of actions to execute. 
Engineered systems commonly address this challenge using modular architectures connected by curated intermediate representations, including, 
for example, a perceptual module for object localization, a grounding module to map localization results to the instruction, and a planner to select the trajectory. 
The required engineering effort is challenging to scale to complex environments.
In this paper, we study a learning-based approach to directly predict continuous control commands given an instruction and visual observations.
This approach offers multiple benefits, including not requiring explicit design of intermediate representations, implementing planning procedures, or separately training multiple sub-models.
We demonstrate the effectiveness of our approach on continuous control of a quadcopter for navigation tasks specified with symbolic instructions.

\begin{figure}

\includegraphics[scale=0.31]{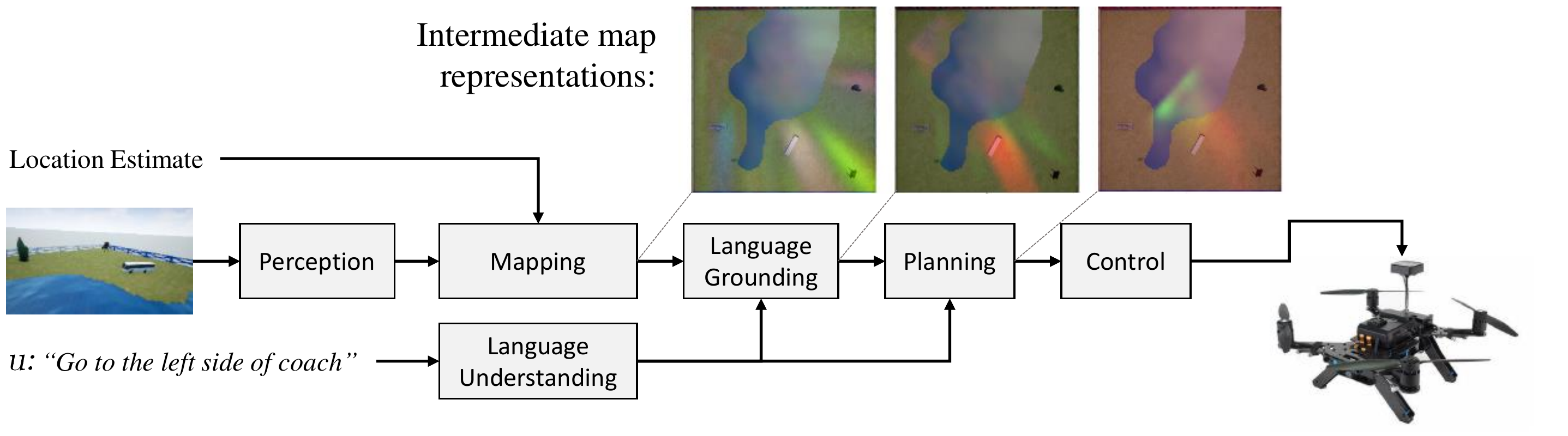}
\caption{A high-level illustration of the $\modelname$ model. Each block represents a neural network or a deterministic differentiable computation. We overlay the different map representations created on an overhead view of the environment to illustrate how the different maps interpret the various environment elements. }
\label{figure:overall_diagam}

\end{figure}

Mapping instructions to actions on a quadcopter requires addressing multiple challenges,
including building an environment representation by reasoning about observations,
recovering the goal from the instruction, 
and continuous control in a realistic environment. 
We address these challenges with the $\fullmodelname$ ($\modelname$) model (\figref{figure:overall_diagam}). The model consists of a single neural network that explicitly maintains a semantic map by projecting learned features from the agent camera frame into the global reference frame.
The map representation is learned from data and can include not only occupancy probabilities, but also high-level semantic information, such as object classes and descriptions. 
The alignment between the map and the environment enables the agent to accumulate memory of features that disappear from the view and 
avoid
the difficulty of reasoning directly about partially-observed first-person observations.

We train the agent to mimic an expert policy using a variant of  $\vanilladagger$~\cite{ross2011reduction}. 
The flexibility of the model makes learning generalizable representations from instructions, observations, and expert actions challenging. 
We use a set of auxiliary objectives to help the different parts of the model specialize as expected. For example, we classify the objects mentioned in the instruction from the intermediate map. 
The auxiliary objectives also solve the credit assignment problem. 
Any failure can be easily attributed to one of the components, but the entire model is still trained end-to-end, which allows later modules to correct previous mistakes.

We evaluate our approach in a simulated quadcopter environment with language instructions generated from a pre-defined set of templates. 
Our model is continuously queried and updated at a rate of 5Hz. 
Our experiments demonstrate that $\modelname$ significantly outperforms standard recurrent architectures that combine convolutional and recurrent layers. 
Our simulator, code, data, and models are available at \href{https://github.com/clic-lab/gsmn}{{\tt https://github.com/clic-lab/gsmn}}.

\section{Technical Overview}
\label{sec:tech_overview}

\newcommand{\configuration}{c}

\subsection{Task}
We model instruction following as a sequential decision-making process~\cite{duvallet2013imitation}. 
Let $\allinstructions$ be the set of instructions, $\allstates$ the set of world states, and $\allactions$ the set of all actions. 
An instruction $\instruction$ is a sequence of $n$ tokens $\langle \instruction_1, \dots, \instruction_n \rangle$. 
Given a start state $\state_1 \in \allstates$ and an instruction $\instruction \in \allinstructions$, the agent executes $\instruction$ by generating a sequence of actions, where the last action is the special action $\stopaction$.
The agent behavior in the environment is determined by its configuration $\configuration$, which specifies the controller setpoints. 
Actions deterministically modify the agent configuration or indicate task completion.
An execution is an $m$-length sequence $\langle (\state_1, \action_1), \dots, (\state_m, \action_m) \rangle$, where $\state_j \in \allstates$ is the  state observed, $\action_j \in \allactions$ is the action updating the agent configuration and $\action_m = \stopaction$. 

In our navigation task, the agent is a quadcopter flying between landmarks in a simulated 3D environment. The state $s$ specifies the full configuration of the simulator, including the positions of all objects and the quadcopter configuration. 
The quadcopter location in the environment is given by its pose $\pose = (\position, \rot)$, where $\position$ is a position and $\rot$ is an orientation. 
The quadcopter has a proportional-integral-derivative (PID) flight controller that maintains a fixed altitude, and takes as input the configuration $\configuration$, which consits of two target velocities: linear velocity $\velfwd \in \reals$ and angular yaw-rate $\velang \in \reals$. 
An action $\action$ is either a tuple $(\velfwd, \velang)$ of velocities or the completion action $\stopaction$. 
Given an action $\action_j = (\velfwd_j, \velang_j)$, we set $\configuration_j = (\velfwd_j, \velang_j)$.  
We observe the environment and generate actions at a fixed rate of 5Hz. The environment simulation runs continuously without interruption. 
Between actions, the quadcopter configuration is maintained. 
To correctly complete a task, the agent must take the $\stopaction$ action at the goal position.

\subsection{Model}
The agent observes the environment via a monocular camera sensor, and has access to its location. 
We distinguish between the world state, which includes the locations of all landmarks and the agent, and the \emph{agent context} $\context$.
The agent has access to the agent context only, including for choosing actions. 
The agent context $\context_j$ at step $j$ is a tuple $(\instruction, \image_j, \pose_j)$, where $\instruction \in \allinstructions$ is an instruction, $\image_j$ is an RGB image, and $\pose_j$ is the agent pose. $\image_j$ and $\pose_j$ are generated from the current world state $\state_j$ using the functions $\imagefunc(\state_j)$ and $\posefunc(\state_j)$ respectively. 
We model the agent using a neural network that explicitly constructs and maintains a semantic map of the environment during execution, and uses the instruction to identify goals in the map. 
At each step $j$, the network takes as input the agent context $\context_j$, and predicts the next action $\action_j$. 
We formally define the agent and model in Section~\ref{sec:model}.

\subsection{Learning}
We assume access to a training set of $N$ examples $\{ (\instruction^{(i)}, \state_1^{(i)},  \posseq^{(i)}) \}_{i = 1}^N$, where $\state_1^{(i)}$ is a start state, $\instruction^{(i)}$ is an instruction, and $\posseq^{(i)} = \langle \pose_1^{(i)}, \dots, \pose_m^{(i)}\rangle$ is a sequence of $m$ poses that defines a trajectory generated by a demonstration execution of $\instruction$. The first pose $\pose_1^{(i)}$ is the  quadcopter pose at state $\state_1^{(i)}$. 
Given $\posseq^{(i)}$, we design an expert oracle policy using a simple path-following carrot planner tuned to the quadcopter dynamics. 
During training, for all states, the oracle policy generates actions that move the quadcopter towards and along the demonstration path. 
We train the agent to mimic the expert policy using a variant of the $\vanilladagger$~\cite{ross2011reduction} algorithm (Section~\ref{sec:learn}).

\subsection{Evaluation}
We evaluate task completion error on a test set of $M$ examples $\{(\instruction^{(i)}, \state^{(i)}_1, \goalpos^{(i)}, \successregion^{(i)}) \}^M_{i=1}$, where $\instruction^{(i)}$ is an instruction, $\state^{(i)}_1$ is a start state, $\goalpos^{(i)}$ is the goal position, and $\successregion^{(i)}$ is the successful completion region defined by an area surrounding $\goalpos^{(i)}$. 
We consider a task as completed correctly if the quadcopter takes the $\stopaction$ action inside $\successregion^{(i)}$ (Section~\ref{sec:experiment}). 

\begin{figure*}[t]
\centering
\includegraphics[scale=0.32]{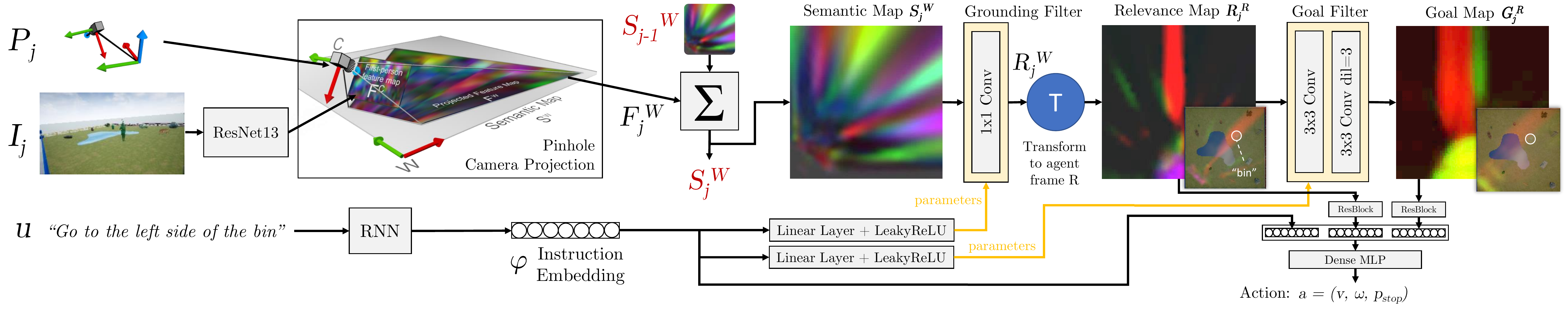}
\caption{An illustration of our model architecture. The instruction $\instruction$ is encoded with an $\lstm$ network into an instruction embedding at the start of an episode. At each timestep, image features are produced with a custom residual network~\cite[ResNet;][]{he2016deep} network and projected on the ground in the global reference frame using a 3D projection based on a pinhole camera model. The projected environment representations are accumulated through time using a masked leaky integration into a persistent map of the world. This map is then filtered to reason about the relevant objects and the most likely goal location using convolutional filters produced from the instruction embedding. From the resulting goal and relevance maps, a dense perceptron (Dense MLP) controller produces a velocity command that drives the robot towards its goal.}
\label{fig:arch}
\end{figure*}
\section{Related Work}
\label{sec:related}

Mapping natural language instructions to actions has been studied extensively, both using  physical robots~\cite{matuszek2012joint, duvallet2013imitation, walter2013learning, tellex11grounding, misra2014context, hemachandra2015learning, knepper15} and virtual agents~\cite{macmahon2006walk, matuszek2010following, matuszek2012learning, artzi2013weakly, branavan2010reading}. 
These approaches are based on a modular system architecture, with separate components for language parsing, grounding, mapping, planning, and control.  
While decomposing the problem, the modular approach requires to explicitly design symbolic intermediate representations, a challenging task for large and complex environments.  
In contrast, we study a single model-approach using a differentiable model architecture that maps visual observations directly to actions, while learning intermediate representations.
This type of approach was studied recently for virtual agents with discrete control~\cite{misra2017mapping, chaplot2017gated, hermann2017grounded, mattersim}. 
In contrast, we study a continuous control problem using a realistic quadcopter simulator. 
We follow existing work~\cite{chaplot2017gated, hermann2017grounded} and abstract the natural language problem by using synthetically generated language. This provides a simple way to specify high-level goals, while focusing our attention on the problem of mapping, planning and task execution.

Recently, using a single differentiable model to map from inputs to outputs across multiple sub-problems has been applied to learning robotic manipulation and control skills~\cite{levine2016learning, levine2016end} and visual navigation in simulated environments~\cite{richter2017safe}.
Similar to recent single model instruction following methods~\cite{misra2017mapping, chaplot2017gated, hermann2017grounded}, these policies are able to learn to effectively complete complex tasks, but suffer from a lack of interpretability. 
In contrast, we design our model to provide an interpretable view of the agent's understanding via the semantic map. 
Our goal is orthogonal to providing safety guarantees~\cite{richter2017safe, held2017probabilistically, kahn2017plato}. 

Key to our approach is building a semantic map of the environment within a neural network model. 
Building environment maps that incorporate information about the semantics of the environment has been studied extensively, commonly with probabilistic graphical models~\cite{persson2007probabilistic, pronobis2011semantic, pronobis2012large, walter2013learning, espinace2013indoor}.
In contrast, our semantic map is a differentiable 3-axis tensor that is part of a larger neural network architecture and stores a feature vector for every observed location in the world. This approach does not require maintaining a distribution over likely maps.
Using a differentiable mapper and planner has been studied for navigation in discrete environments~\citep{gupta2017cognitive, gupta2017unifying, parisotto2018, khan2018memory}. 
We work with continuous action and state spaces, and emphasize efficient learning from limited data by incorporating explicit projection and simple aggregation instead of learned memory operations. 
Including affine transformations and projections inside a neural network has been previously studied in vision and graphics~\cite{jaderberg2015spatial, yan2016perspective}.  We use these techniques for learning map-based environment representations.

We evaluate our approach on a realistic simulated quadcopter performing a high-level navigation task.
Quadcopters have been recently studied with the goal of learning low-level continuous  control~\cite{mueller2017teaching,ross2013learning} or navigation policies~\cite{sadeghi2016cad2rl, giusti2016machine}, where navigation was cast as traversable space prediction using supervised learning. 
In contrast, we focus on mapping high-level symbolic instructions directly to control signals.
For learning, we use imitation learning~\cite{bakker1996robot, schaal2003computational, daume09searn}, where an agent policy is trained to mimick an expert policy while learning how to recover from errors not present in the expert demonstrations. 
We use a variant of \textsc{DAgger}~\cite{ross2011reduction}, where states and actions are aggregated from an expert policy  for supervised learning. 
\citet{hussein2017imitation} provides a general overview of imitation learning.

\section{Model Architecture}
\label{sec:model}

We model the agent policy $\policy$ with a neural network. 
At time $j$, the input to the policy is the agent context $\context_j$, and the output is an action $\action_j$. The action $\action_j$ modifies the agent configuration $\configuration_j$. This process continues until the $\stopaction$ action is predicted and the agent stops. The agent context $\context_j$ is a tuple $(\instruction, \image_j, \pose_j)$, where $\instruction$ is the instruction, $\image_j = \imagefunc(\state_j)$ is the current observation, and $\pose_j = \posefunc(\state_j)$ is the pose of the agent.
\figref{fig:arch} illustrates our network architecture.

Our model design incorporates explicit spatial reasoning and memory operations into a differentiable neural network. This relieves the neural network from learning to accomplish complex coordinate transformations that map between the camera and the world reference frame and from learning to integrate current and past observations into a coherent world model. 
We incorporate a 3D projection and a coordinate frame transformation into our image processing pipeline. 
Feature representations seamlessly propagate through these operations in a differentiable manner, while the transformations themselves are not learned. 
The projected image representation is added into a persistent semantic map defined in the global reference frame and aligned with the environment, which allows the agent to easily retain information accumulated through time. 
This map is similar in principle to the way simultaneous localization and mapping (SLAM) systems store low-level features, such as LIDAR bounces, depth readings or feature descriptors.
However, unlike SLAM maps, the features stored are representations learned by our differentiable mapper to directly optimize the task performance.

\subsection{Instruction and Image Embedding}

We generate representations for both the input text $\instruction$ and the image $\image$. 
We use a recurrent neural network~\cite[RNN;][]{Elman:90rnn} with a long short-term memory recurrence~\cite[LSTM;][]{hochreiter1997long} to generate a sentence embedding of the input instruction text $\embed_\instruction$ by taking the last $\lstm$ output.

\begin{figure}[t]
\centering
\includegraphics[scale=0.3]{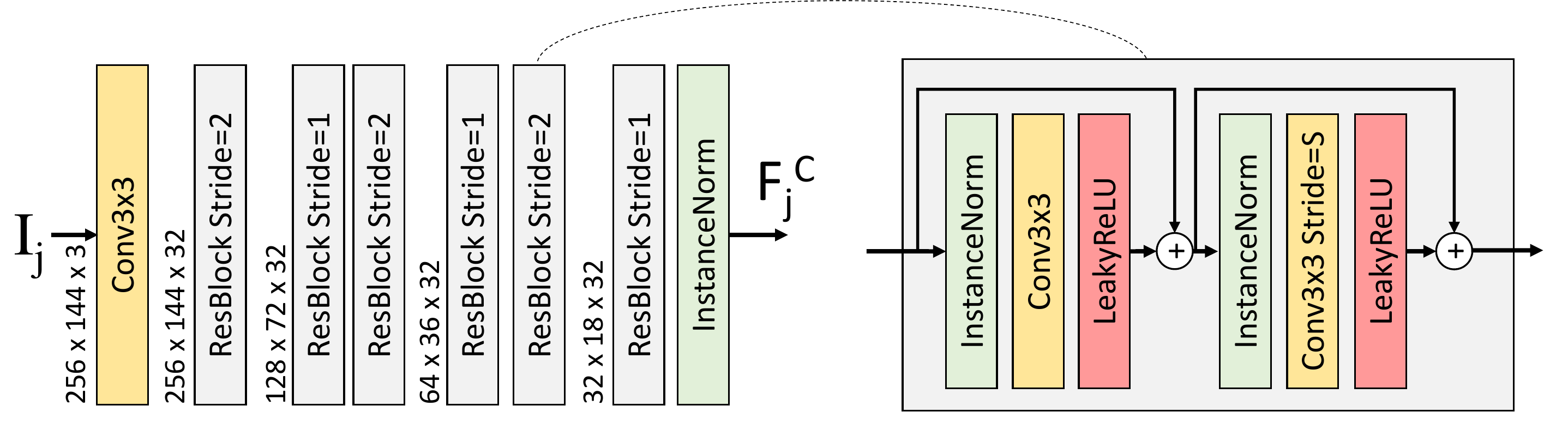}
\caption{Illustration of our ResNet architecture. The ResNet architecture (left) contains six residual blocks (right). The ResNet has 13 convolutional layers: each block has two convolutional layers and there is one layer at the ResNet input.}
\label{fig:resnet}
\end{figure}

Given the observed image $\image_j$, we compute a feature representation $\fmcam_j = \resnet(\image_j)$ using a 13-layer residual convolutional neural network~\cite{he2016deep} (\figref{fig:resnet}). 
For a given image $\image_j$ of size $H\text{x}W\text{x}3$, $\fmcam_j$ is a feature map of size $(H/f_{\text{scale}})\text{x}(W/f_{\text{scale}})\text{x}C_f$, where the factor $f_{\text{scale}}$ is a hyperparameter of the  $\resnet$. Each pixel in $\fmcam_j$ is a feature-vector of $C_f$ elements (i.e., channels) that encodes properties of the corresponding image region receptive field. Each pixel in $\fmcam_j$ embeds a  different image region, which allows recovering locations of objects visible in $\image_j$ given $\fmcam_j$.
In our implementation, $C_f = 32$, $f_{\text{scale}}$ = 8 and the receptive field of each feature vector in $\fmcam_j$ is a 61x61 neighborhood in the image $\image_j$. The resolution of $\image_j$ is 256x144.

\subsection{Feature Projection}
The image representation $\fmcam_j$ is oriented in the first-person view corresponding to the camera image plane.
We project $\fmcam_j$ on the ground in the world reference frame to align with the environment and integrate it with previous observations.
The current location of each element feature vector $i$ in $\fmcam_j$ is represented in homogeneous 2D pixel coordinates as $p^\camframe_i = [x^\camframe_i, y^\camframe_i, 1]^T$, where $x^\camframe_i$ and $y^\camframe_i$ are the 2D coordinates of $i$ in $\fmcam_j$.
We use a pinhole camera model to project each element $i$ to the ground in the world reference frame:
\begin{equation}
\nonumber p^\worldframe_i = T_{\camframe}^{\worldframe}\raycast(K^{-1}p^\camframe_i)\mbox{,}
\end{equation}
where $K$ is the camera matrix,  $T_{\camframe}^{\worldframe}$ is a camera-to-world affine transformation deterministically computed from the agent pose $\pose_j$, and $\raycast$ is a function that maps a ray to the point where the ray intersects the ground plane at elevation $0$. 
\citet{poling2015tutorial} provides a brief tutorial of camera models. 
We construct the observation representation $\fmworld_j$ in the world reference frame by copying each element $i$ in $\fmcam_j$ to the location $p^\worldframe_i$ in $\fmworld_j$. 
To generate the final $\fmworld_j$, for each pixel, we  interpolate the neighboring points by applying bi-linear interpolation~\cite{jaderberg2015spatial}. This creates a discretized  tensor representation of $\fmworld_j$ and resolve cases where multiple elements are placed in the same pixel location.

\subsection{Semantic Map Accumulation}
We use the transformed feature map $\fmworld_j$ to update a persistent semantic map $\smworld_j$, where we accumulate visual information through time. 
The map $\smworld_j$ is a 3D tensor with two spatial dimensions and one feature vector dimension to store the different channels generated by the $\resnet$. 
Given the feature map $\fmworld_j$ computed at step $j$ and the semantic map $\smworld_{j-1}$ from step $j-1$, we compute the semantic map $\smworld_j$ at step $j$ with a leaky integrator filter: 
\begin{eqnarray}
& \nonumber \smworld_j = (1 - \lambda)\smworld_{j-1} + \lambda(\fmworld_j \odot \maskworld_j) + \\ & \hspace{3.85cm} \lambda(\smworld_{j-1} \odot (1 - \maskworld_j))\mbox{,}	\nonumber
\end{eqnarray}
where $\maskworld_j$ is a binary-valued mask indicating which pixels in $\smworld_j$ are within the agent's field of view and $\odot$ is an element-wise multiplication.
This update ensures that (a) unobserved locations (e.g., behind the robot, outside the view of the camera) are not updated, and (b) information is combined with what is already in the map so that an erroneous reading does not fully overwrite valid prior information. The model is able to tolerate moderate amounts of noise in the pose estimate, a likely source of noise in physical robots with GPS or SLAM based localization. We test noise-tolerance in \secref{sec:results}.

\subsection{Language Grounding and Goal Prediction}
We use the instruction embedding $\embed_\instruction$ to create two maps from the semantic map: a relevance map that accounts for landmarks mentioned in the instruction and a goal map to identify the goal location. 
To compute the relevance map, we use the instruction embedding to create a language-dependent 1x1 convolutional filter: 
\begin{equation}
\nonumber \convlab_{\lab} = \weights_{\lab}\embed_\instruction + \bias_{\lab}\mbox{,}
\end{equation}
where $\embed_\instruction$ is the instruction embedding and $\weights_{\lab}$ and $\bias_{\lab}$ are learned parameters. 
The relevance map $\rmworld$ aims to identify the objects in the semantic map  mentioned in $\instruction$:\footnote{The 1x1 convolution operation is equivalent to an affine transformation performed on each value (i.e., feature vector) in $\smworld_j$. This allows the output map to store semantic information conditioned on the instruction.}
\begin{equation}
\nonumber	\rmworld_j = \convlab_{{\lab}}(\smworld_j)\mbox{.}
\end{equation}
The goal map $\gmrobot$ is computed from the relevance map with wider convolutions to capture spatial relationships.
While the relevance map is computed in the global world reference frame, instructions are usually given in the ego-centric agent reference frame. Before computing the goal map, we compute $\rmrobot_j$ by transforming $\rmworld_j$ to the agent reference frame. 
We use a separate convolution filter computed from the instruction $\instruction$: 
\begin{equation}
\nonumber \convspat_{\spat} = \weights_{\spat}\embed_\instruction + \bias_{\spat}\;\;,
\end{equation}
where $\weights_{\spat}$ and $\bias_{\spat}$ are learned parameters.
$\convspat_{\spat}$ consists of two cascaded 3x3 convolutions.\footnote{Both convolutions use a kernel size of $3$ and $\leakyrelu$ activations. The second convolution uses a dilated kernel~\cite{YuKoltun2016} with dilation of 3 to increase the receptive field of the filter.}
The goal map is computed as 
\begin{equation}
\nonumber	\gmrobot_j = \convspat_{\spat}(\rmrobot_j)\;\;.
\end{equation}

\subsection{Control}
To compute the output action $\action_j$, we use a densely-connected two-layer perceptron. Densely connected neural networks have been shown as more stable and faster to train than standard feed-forward models~\cite{huang2017densely}. The input to the perceptron is a concatenation of pre-processed relevance and goal maps and the instruction embedding:
\begin{equation}
\embed_{p_{in}} = [\resblock_{R}(\rmrobot_j), \resblock_{G}(\gmrobot_j), \embed_{\instruction}]\;\;,\nonumber
\end{equation}
where $\resblock_{(\cdot)}$ is a residual block (\figref{fig:resnet}) with a stride of two to reduce the spatial dimensionality of the maps.
Formally, the densely-connected perceptron is: 
\begin{eqnarray}
\nonumber \embed_{p_1} &=& l(W_{p_1}\embed_{p_{in}} + b_{p_1}) \quad \\
\nonumber \action_j &=& W_{p_2}[\embed_{p_{in}}, \embed_{p_1}] + b_{p_2}\mbox{,}
\end{eqnarray}
where $l(\cdot)$ is a $\leakyrelu$ activation function~\cite{maas2013rectifier} and $\action_j$ is the output action of the policy $\policy$.

\subsection{Initial Values and Parameters}

At the beginning of execution every element of $\smworld_0$ is set to $\vec{0}$. The model parameters $\param$ include the word embeddings used as input for the $\lstm$, the $\lstm$, the $\resnet$, $\resblock_R$, $\resblock_G$ and the matrices and bias vectors: $\weights_{\lab}$, $\weights_{\spat}$, $W_{p_1}$, $W_{p_2}$, $\bias_{\lab}$, $\bias_{\spat}$, $b_{p_1}$, $b_{p_2}$. 
The map transformations and observation projections are computed deterministically given the agent pose.

\section{Learning}
\label{sec:learn}

We estimate the parameters of the model $\param$ using imitation learning with $\daggerfm$, a variant of $\vanilladagger$~\cite{ross2011reduction} for memory-limited training scenarios.\footnote{FM in $\daggerfm$ stands for fixed memory.}
While $\vanilladagger$ provides realistic sample complexity and has been shown to work on robotic agents~\cite{ross2013learning}, it requires maintaining an ever-growing dataset to provide stable training. In complex continuous visuo-motor control settings, such as ours, the number of samples generated is large and each sample requires significant amount of memory. This quickly results in exhaustion of memory resources. 
$\daggerfm$ trades the guarantees of $\vanilladagger$ for a fixed memory budget.

We assume access to a training data of $N$ examples $\{ (\state_1^{(i)}, \instruction^{(i)}, \posseq^{(i)}) \}_{i = 1}^N$, where $\state_1^{(i)}$ is a start state, $\instruction^{(i)}$ is an instruction, and $\posseq^{(i)} = \langle \pose_1, \dots, \pose_m\rangle$ is a demonstration sequence of $m$ poses starting at $\pose_1^{(i)} = \posefunc (\state_1^{(i)})$. 
Given an expert policy $\oracle$ and a training example $(\state_1^{(i)}, \instruction^{(i)}, \posseq^{(i)})$, we minimize the expected distance between the policy actions $\action = \policy(\context)$ and the expert action $\expertaction=\oracle(\pose, \posseq^{(i)})$. The expectation is computed over the state distribution $\policystatedist$ induced when executing the learned policy $\policy$ starting from $\state_1^{(i)}$ and using $\instruction^{(i)}$:
\begin{equation}
J(\param) = \mathbb{E}_{\state_j \sim \policystatedist}[\actiondistance(\action, \expertaction)]\mbox{.}
\label{eq:action_objective}
\end{equation}
During learning, we sample states $\state_j$ from the state distribution $\policysampledist$ induced by the mixture policy $\hat{\pi}$, which converges to $\policystatedist$
The distance metric $\actiondistance$ is defined as:
\begin{eqnarray}
\nonumber  \actiondistance(\action, \expertaction) =& \hspace{-2.85cm} \|\velfwd - \velfwdstar\|^2_2 + \|\velang - \velangstar\|^2_2 + \\
\nonumber &  [\stopprobstar log(\stopprob) + (1 - \stopprobstar) log (1 - \stopprob)]\;\;,
\end{eqnarray}
where $\action = (\velfwd, \velang, \stopprob)$ and $\expertaction = (\velfwdstar, \velangstar, \stopprobstar)$. The policy outputs the $\stopaction$ action when $\stopprob > 0.5$.

Algorithm~\ref{alg:full} shows the training algorithm. 
Learning begins by collecting a dataset $\supdataset$ of $N_s$ trajectories using the expert policy $\oracle$ for supervised learning. 
Following supervised training on $\supdataset$, we sample the initial dataset $\mathcal{D}$ from $\supdataset$. 
We then iterate for $K$ iterations. At each iteration, we discard $N_d$ trajectories from $\mathcal{D}$, collect new $N_d$ trajectories using a mixture policy that interpolates the oracle $\oracle$ and the current policy $\policy_i$, update the dataset with the new trajectories, and do one epoch of gradient updates with the aggregated dataset.

$\daggerfm$ does not provide a convergence guarantee for a tabular policy. However, we empirically observe that the dataset performs a stabilizing function similar to that of replay memory in deep Q-learning~\cite{mnih2015human}.

\begin{algorithm}[t]
\begin{algorithmic}
\State $\mathcal{D^*} \leftarrow \texttt{collect\_dataset} (\pi^*, N_s)$
\State $\pi_{\theta_1} \leftarrow \texttt{train\_supervised}(\mathcal{D^*})$
\State Sample initial dataset $\mathcal{D} \sim \mathcal{D^*}$ of size $N$
\For {$i = 1$ to $K$}
	\State Discard $N_d$ trajectories from $\mathcal{D}$ uniformly at random
	\State Decay $\beta$: $\beta \leftarrow (\beta_0)^{i}$
	\State Let $\hat{\pi}_i = \beta\pi^* + (1 - \beta)\pi_i$
	\State $\mathcal{D}_i \leftarrow \texttt{collect\_dataset} (\hat{\pi}_i, N_d)$ of size $N_d$
	\State $\mathcal{D} \leftarrow \mathcal{D} \cup \mathcal{D}_i$
	\State $\pi_{i+1} \leftarrow \texttt{train\_epoch}(\mathcal{D}, \pi_i)$
\EndFor
\Return $\pi_K$

\end{algorithmic}
\caption{DAggerFM Training Algorithm for imitation learning with capped dataset size.}
\label{alg:full}
\end{algorithm}

\section{Auxiliary Objectives}
\label{sec:objectives:main}

The decomposition of the model architecture according to the different types of  expected reasoning (i.e., perception, grounding, and planning) allows us to easily define appropriate auxiliary objectives. 
These objectives encourage the different parts of the model to assume their intended functions and solve the credit assignment problem. 
We add four additive auxiliary objectives to the training objective: 
\begin{eqnarray}
\nonumber	J(\param) &=& \jaction(\param) + \weightclass \jclass(\param) + \\
\nonumber	&& \weightlang \jlang(\param) + \weightrel \jrel(\param) + \weightgoal \jgoal(\param)\;\;,
\end{eqnarray}
where $\jaction(\cdot)$ is the main objective (Equation~\ref{eq:action_objective}) and the four auxiliary objectives are $\jclass(\cdot)$, $\jlang(\cdot)$, $\jrel(\cdot)$, and $\jgoal(\cdot)$. Each auxiliary objective is weighted by a coefficient hyper-parameter. 

The perception objective $\jclass(\cdot)$ aims to require the perception components of the model to correctly classify visible objects. 
At time $j$, for every object $o$ visible in $\image_j$, we classify the element in the semantic map $\smworld_j$ corresponding to its location in the world.
We apply a linear softmax classifier to every semantic map element that spatially corresponds to the center of an object. The classifier loss is: 

\begin{equation}
\nonumber \jclass(\param) = \frac{-1}{|O_{{\rm FPV}}|}\sum_{o \in O_{{\rm FPV}}}[\hat{y}_o log(y_o)]\;\;,
\end{equation}
where $\hat{y}_o$ is the true class label of the object $o$ and $y_o$ is the predicted probability.  The instruction objective $\jlang(\cdot)$ defines a similar objective for the instruction representation. Given an instruction, it requires to classify the object mentioned and the side of the goal (e.g., right, left). The objective is otherwise identical to $\jclass(\cdot)$. 

The grounding and planning objectives $\jrel(\cdot)$ and $\jgoal(\cdot)$ use binary classification on positions in the relevance and goal maps. 
Both objectives use a binary cross-entropy loss. 
The relevance map objective $\jrel(\cdot)$ classifies each object on the relevance map as to whether it was mentioned in the instruction $\instruction$ or not. For each timestep $j$, we average the objective over all objects in the agent's field of view. The object mentioned in the instruction is a positive example, and all others are negative examples. 
The objective is: 
\begin{equation}
\nonumber \jrel(\param) = \frac{-1}{|O_{{\rm FPV}}|}\sum_{o \in O_{{\rm FPV}}}[\hat{y}^r_o log(y^r_o) + (1 - \hat{y}^r_o)log(1 - y^r_o)]\mbox{,}
\end{equation}
where $O_{{\rm FPV}}$ is the set of objects visible from the agent perspective,  $\hat{y}^r_o$ indicates if the object was mentioned in the instruction $\instruction$, and $y^r_o$ is the prediction of a linear classifier.
The goal map objective $\jgoal(\cdot)$ is a similar binary objective, and classifies a point on the goal map as being the goal or not. Positive examples are taken form the annotated goals. For each goal, a negative example is randomly sampled.

\section{Experimental Setup}
\label{sec:experiment}

\begin{figure}[t]
\centering
\includegraphics[width=\linewidth]{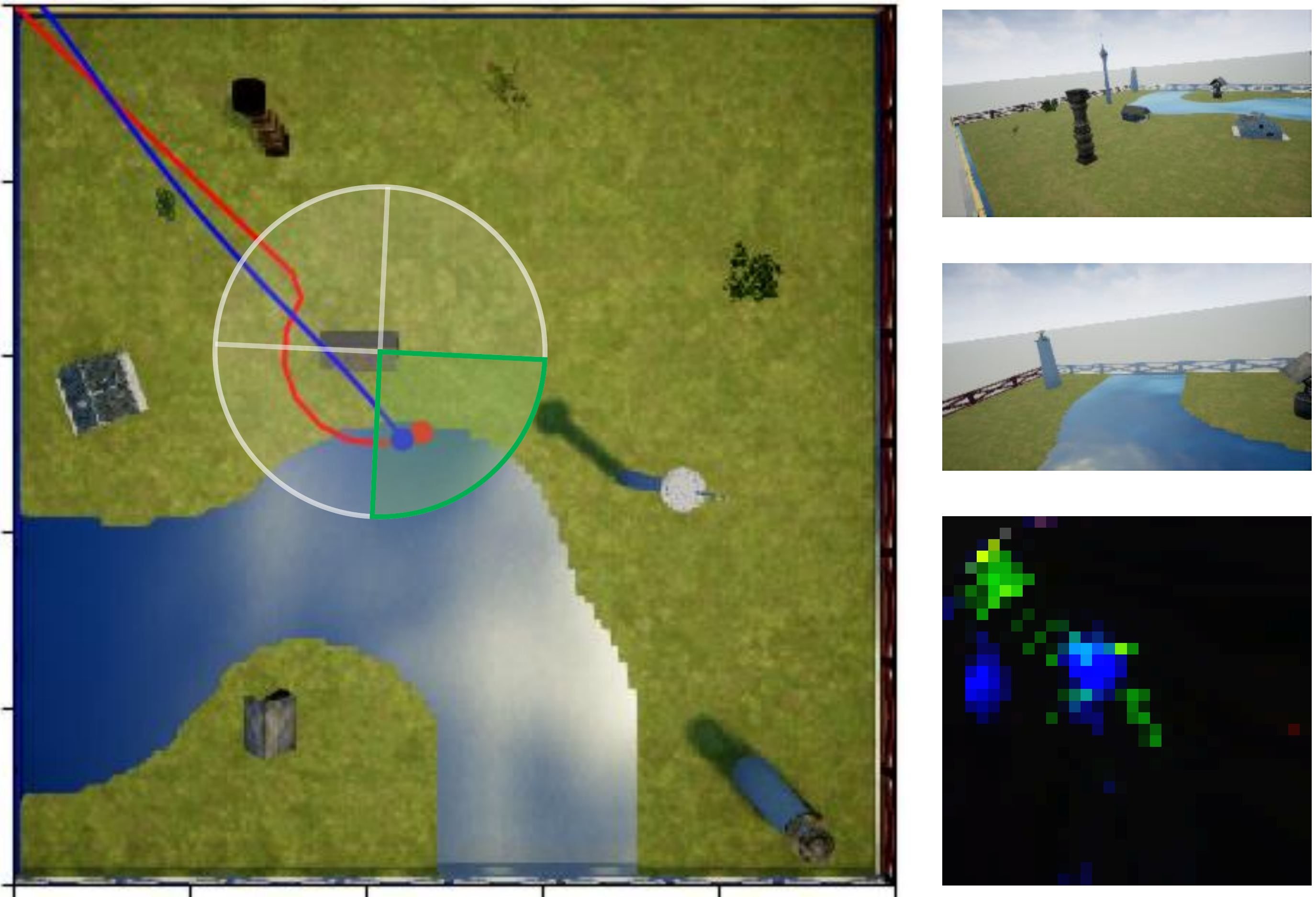}
\fbox{
\textit{Go to the back side of wooden house}}
  \caption{Example environment (left) with 9 objects and an instruction. Ground truth trajectory is shown in red. The trajectory taken by our learned agent policy is in blue. The shaded circle is the target landmark region with $\lmpos - \pos_m < \stopthres$, and the quadrant highlighted in green is the successful completion region. On the right, from the top: the first observation $\image_0$, the last observation $\image_{37}$, and the final relevance map $\rmworld_{37}$ produced by $\modelname$ model. The relevance map $\rmworld_{37}$ shows the instruction grounding result with both houses highlighted in blue and the wooden house receiving the stronger highlight. The agent correctly stops in the completion region even though the house is outside its field of view in $\image_{37}$.}
\label{fig:example_env}
\end{figure}

\subsection{Environment}
We evaluate our approach in randomly generated virtual environments in Unreal Engine. \figref{fig:example_env} shows an example environment.
Each environment consists of a square-shaped green field with edge length of 30 meters and $\numlandmarks$ landmarks.
We choose $\numlandmarks$ uniformly at random between 6 and 13 inclusive and draw landmarks from a pool of 63  3D models without replacement.
Landmarks are placed in random locations but no closer than 2.7 meters to the environment edges and at least 1.8 meters apart.
We fill the remaining area with random decorative lakes.
The quadcopter starts in one of the four corners of the field, facing inwards with a full view of the field.
This allows us to validate the model capabilities independently of an exploration strategy.

\subsection{Data}
For each environment, we randomly sample a pair $\gamma = (\gamma^C, \gamma^S)$, where $\gamma^C$ is one of the 63 landmarks and $\gamma^S$ is one of $\{\leftside, \rightside, \frontside, \backside\}$.
Given $\gamma$, we generate the instruction $u$, of the form \textit{"go to the $S$ side of $C$"}, and a goal location $\goalpos$.
There are 45 unique noun phrases for the 63 landmarks, and some instructions are ambiguous and unsolvable by even a perfect model.
The goal location $\goalpos$ is placed between 2.25 and 3.75 meters from the landmark depending on its size, on the side corresponding to $\gamma_S$ when viewed from the start position $\position_0$. We generate the ground-truth demonstration trajectory $\posseq$ by simulating a point-mass attracted to $\goalpos$ with landmark avoidance constraints.
We generate a total of $\trainsetsize$ training, $\devsetsize$ development and $\testsetsize$ testing environments.
The same 63 landmarks are found in all dataset splits, but in different combinations and locations.
We additionally report results on a pruned test set, where environments with ambiguous instructions have been excluded.

\begin{table*}[t!]
\begin{center}
\begin{tabular}{lccccc}
 & \multicolumn{2}{c}{Success Rate (\%)} & \multicolumn{2}{c}{Distance to Goal (m)} & \\
 & Overall & Landmark & Mean & Median  &  \\
\midrule
\textbf{\modelname (Ours)}     & \best{83.47} (\best{87.21})    & \best{89.33} (\best{93.38})    & \best{2.67} (\best{2.24})    & \best{1.21} (\best{1.12})   &  \\
\modelnogoal                   & 69.89 (71.03)    & 84.76 (89.56)    & 3.19 (2.91)    & 1.63 (1.65)   &  \\
\midrule
\blimg                         & 24.93 (27.35)    & 56.80 (56.03)    & 7.23 (7.15)    & 5.14 (4.97)    &  \\
\blimgrec  			           & 28.67 (33.82)    & 60.13 (65.74)    & 6.72 (6.11)    & 4.34 (3.95)     &  \\
\midrule
\midrule
Oracle (Expert)                & 87.87 (86.47) & 98.40 (98.10) & 0.74 (0.78)  & 0.67 (0.73)   &  \\
Avg \# Steps Fwd               & 3.32 (3.98)   & 49.00 (14.12) & 11.96 (13.43) & 12.23 (13.71)  &  \\
Random Point                   & 2.13         & 9.59        & 15.14       & 15.05       &  \\
Random Landmark                & 17.31        & 17.31        & 13.26       & 13.37       &  \\
\midrule
\end{tabular}
\end{center}

\captionsetup{justification=justified}
\caption{Test results. The numbers in brackets show performance on a pruned version of the test set containing only the $\prunedtestsetsize$ environments that do not include ambiguous instructions.
}
\label{tab:test_results}
\end{table*}

\subsection{Evaluation Metric}

Given the  instruction template $\gamma$, let $\lmpos$ be the location of the target landmark $\gamma^C$ in the environment. We define the target landmark region as the circular area where the Euclidean distance of the agent's last position $\pos_m$ is less than a threshold $\stopthres$ from $\lmpos$. We subdivide the target landmark region into 4 quadrants, corresponding to $\leftside$, $\rightside$, $\frontside$ and $\backside$, with respect to the agent's starting position. We define the task as successfully completed if the agent outputs the action $\stopaction$ within the correct quadrant $\successregion$ of the target landmark region (see \figref{fig:example_env}). 
We use $\stopthres = \hyperstopthres$.

\begin{table}[t!]
\begin{center}
\begin{tabular}{lccccc}
 & \multicolumn{2}{c}{Success Rate (\%)} & \multicolumn{2}{c}{Dst. to Goal (m)} & \\
 & Overall & Landmark & Mean & Median  &  \\
\midrule
\textbf{\modelname (Ours)}     & \best{79.20}        & \best{86.00} & \best{2.98}        & \best{1.21}        &  \\
\modelname - $\jgoal$          & 64.40        & 82.00        & 3.61        & 1.75        &  \\
\modelname - $\jrel$           & 5.73         & 17.73        & 10.69       & 10.64       &  \\
\modelname - $\jlang$          & 41.20        & 57.20        & 6.62        & 3.37        &  \\
\modelname - $\jclass$         & 50.00        & 63.73        & 5.89        & 2.55        &  \\
\midrule
\modelname + $\eta_{pos}$      & 74.40        & 85.33        & 3.01        & 1.41        &  \\
\midrule
\end{tabular}
\end{center}

\captionsetup{justification=justified}
\caption{Development results. We compare our approach against ablations of the  different auxiliary objectives and a model that observes noisy poses \modelname + $\eta_{pos}$.}
\label{tab:dev_results}
\vspace{-10pt}
\end{table}

We additionally report the mean and median of the distance between $\pos_m$ and the ground truth goal position $\goalpos$, as well as the fraction of executions in which the agent stopped in the target landmark region, but on any side of the landmark.

\subsection{Systems}
We demonstrate the performance of our model $\modelname$ against two  neural network baselines that use first-person view: $\blimgrec$ and $\blimg$.  
$\blimgrec$ processes the feature map $\fmcam$ in the first-person view using the same language-derived filters $\convlab_{\lab}$ and $\convspat_{\spat}$ and applies the same auxiliary objectives to all intermediate representations, resulting in per-timestep relevance and goal maps $\rmcam$ and $\gmcam$ in the first-person view.
The only difference is that no 3D projection takes place and consequently no environment map is built.
Since $\modelname$ uses the semantic map as spatial memory, $\blimgrec$ uses an $\lstm$ memory cell for this purpose~\cite{hermann2017grounded, chaplot2017gated}.
We store the current $\rmcam$ and $\gmcam$ in the $\lstm$ cell and take the current output of the $\lstm$ as an additional input to the $\mlp$.
$\blimg$ is a baseline with the same architecture as $\blimgrec$, but without the $\lstm$ memory.
We do not use $\jgoal$ with the baselines because the goal location may not be visible in view. 
The direct comparison should be made against the $\modelnogoal$ ablation, which uses the  the same set of auxiliary objectives.

We use $\blimg$ and $\blimgrec$ as baselines to clearly quantify the  contribution of the semantic mapping mechanism. 
We also compare against three trivial baseline models: (a) take the average action for average number of steps ($\avgnumsteps$); (b) go to a random point in the field; and (c) go to the correct side of a random landmark in the field. 
As an upper bound, we compare against the expert policy.

Finally, we study the sensitivity of our model to simple noise in the position estimate by adding Guassian noise. 
We consider the pose at time $j$ as a 3D position and 3D Euler angles $\pose_j = [\pos_x, \pos_y, \pos_z, \rot_x, \rot_y, \rot_z]$.
At each timestep $j$, we replace the pose $\pose_j$ with $\hat{\pose_j} = \pose_j + \eta_j$, where $\eta_j = [\eta_{\pos_x}, \eta_{\pos_y}, \eta_{\pos_z}, 0, 0, 0]$ is additive noise. 
Each noise component $\eta_{\pos_{(\cdot)}}$ is drawn at every timestep $j$ independently at random from a Gaussian distribution with zero mean and variance of 0.5 meters.

\subsection{Hyper-parameters and Implementation Details}

We train all neural network models using the same procedure.
We collect a supervised dataset $\supdataset$ consisting of sequences of observations and actions on all $\trainsetsize$ training environments by executing the expert policy.
We then train on $\supdataset$ for 30 epochs, and execute $\daggerfm$ for $100$ epochs with $N_D=520$ and $M=20$ (\secref{sec:learn}).
We optimize the parameters $\param$ of the policy $\policy$ using  $\adam$~\cite{kingma2014adam}  with $\alpha=0.001$, learning rate of $0.001$, L2 regularization with $\gamma=10^{-6}$, and the gradient $\frac{\partial J(\policy)}{\param}$.
We set $\weightclass$, $\weightrel$, $\weightlang$, and $\weightgoal$ to $\lambdaaux$. 
We execute the trained policies and baselines on a simulated quadcopter using the Microsoft AirSim~\cite{airsim2017fsr} plugin for Unreal Engine 4.18, which captures realistic flight dynamics.
We control the quadcopter by sending velocity commands to the flight controller in its local reference frame, and limit the linear velocity to $\maxvelx$ and the angular velocity to $\maxvelw$.

\begin{figure*}[t]
\includegraphics[width=\linewidth]{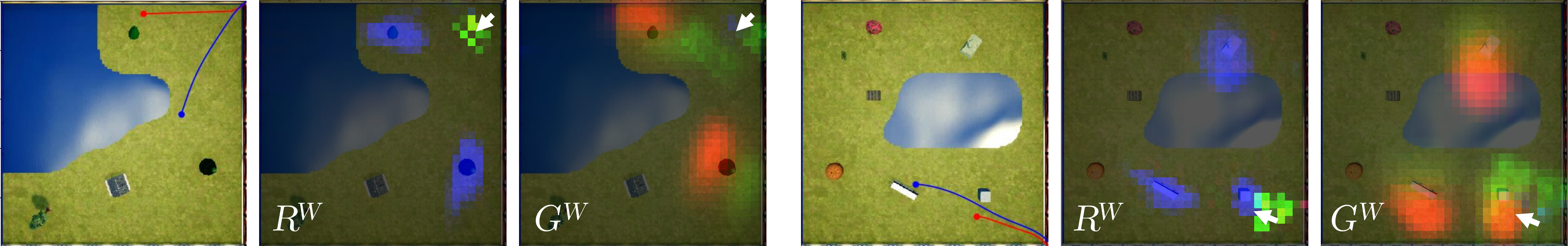}
\hspace*{0.11\textwidth}
\fbox{\textit{Go to the right side of fir tree}}
\hfill
\fbox{\textit{Go to the left side of box}}
\hspace*{0.13\textwidth}
  \caption{Two task failures using our approach. The ground truth and policy trajectories are shown in red and blue respectively. On the left, the failure is due to an ambiguous instruction. The Relevance Map $\rmworld$ overlaid on the environment reveals that the agent has correctly grounded the instruction to the two fir trees present, while ignoring the other objects. The goal map $\gmworld$ shows that the goal location is inferred for both fir trees. The agent correctly reasons about the task, but the ambiguity confuses it. On the right, box is incorrectly grounded to multiple objects as seen in $\rmworld$. Given this wrong grounding, the goal location is inferred to the correct side (left) of all grounded objects, as shown in $\gmworld$. The agent then executes given the confused goal map, which causes it to fly through the correct goal and towards one of the wrongly detected goals.}
\label{fig:fail_analysis}
\end{figure*}

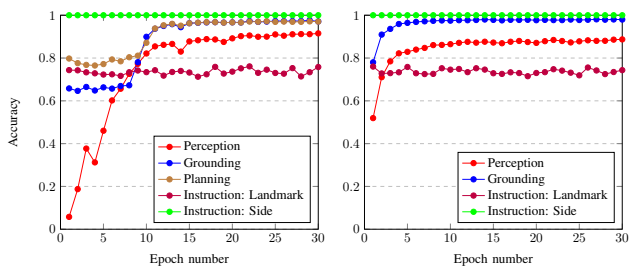
\begin{figure}[t]
\begin{tikzpicture}[yscale=0.5, xscale=0.5]
\begin{axis}[
    xlabel={Epoch number},
    ylabel={Accuracy},
    xmin=0, xmax=30,
    ymin=0, ymax=1,
    legend pos=south east,
    legend cell align=left,
    ymajorgrids=true,
    grid style=dashed,
]
\addplot[red, mark=*]
table[x=index, y=class_acc, col sep=comma]
{data/gsmn.csv};
\addplot[blue, mark=*]
table[x=index, y=gnd_acc, col sep=comma]
{data/gsmn.csv};
\addplot[brown, mark=*]
table[x=index, y=goal_acc, col sep=comma]
{data/gsmn.csv};
\addplot[purple, mark=*]
table[x=index, y=lm_acc, col sep=comma]
{data/gsmn.csv};
\addplot[green, mark=*]
table[x=index, y=side_acc, col sep=comma]
{data/gsmn.csv};
\legend{Perception, Grounding, Planning, Instruction: Landmark, Instruction: Side}
\end{axis}
\end{tikzpicture}
\begin{tikzpicture}[yscale=0.5, xscale=0.5]
\begin{axis}[
    xlabel={Epoch number},
    xmin=0, xmax=30,
    ymin=0, ymax=1,
    legend pos=south east,
    legend cell align=left,
    ymajorgrids=true,
    grid style=dashed,
]
\addplot[red, mark=*]
table[x=index, y=class_acc, col sep=comma]
{data/gs_fpv_mem.csv};
\addplot[blue, mark=*]
table[x=index, y=gnd_acc, col sep=comma]
{data/gs_fpv_mem.csv};
\addplot[purple, mark=*]
table[x=index, y=lm_acc, col sep=comma]
{data/gs_fpv_mem.csv};
\addplot[green, mark=*]
table[x=index, y=side_acc, col sep=comma]
{data/gs_fpv_mem.csv};
\legend{Perception, Grounding, Instruction: Landmark, Instruction: Side}
\end{axis}
\end{tikzpicture}
\caption{Accuracy curves for main objective ($\jaction(\cdot)$) and the different auxiliary objectives: perception ($\jclass(\cdot)$), instruction understanding ($\jlang(\cdot)$), grounding ($\jrel(\cdot)$), and planning ($\jrel(\cdot)$). The curves show the progress of learning in terms of individual accuracies during supervised pre-training measured on the development set. We show the curves for $\modelname$ (left) and $\blimgrec$ (right).}
\label{fig:acc_curves}
\vspace{-10pt}
\end{figure}

\section{Results}
\label{sec:results}

\tabref{tab:test_results} shows our test results. 
Our approach significantly outperforms the baselines. 
While our full approach includes the goal auxiliary objectives, which the baselines do not have access to, we observe that $\modelnogoal$, which ablates this auxiliary objective, also outperforms the RNN baseline $\blimgrec$. 
Our full model $\modelname$ performs very closely to the expert policy, especially when removing ambiguous instructions.
In general, the oracle performs imperfectly due to the simple model we use. For example, at times, the agent reaches the goal position in high speed and ends up stopping just outside the correct completion boundary.

\tabref{tab:dev_results} shows development results. 
We ablate each auxiliary objective. Each objective contributes to the model performance. 
We observe that the grounding objective $\jrel$ is  essential and without it the model fails to learn.
The planning auxiliary objective $\jgoal$ is the least important. This suggests that after having successful grounding results, the set of simple spatial relations we use is relatively easy to learn. 
We also observe that our model is robust to moderate amount of localization noise without a significant decline in performance.
This indicates the model potential for physical robotic systems, where position estimates are likely to be noisy. 

\figref{fig:acc_curves} shows  development accuracy of auxiliary objectives during training as function of the number of epochs. 
We observe that most auxiliary objectives converge for both the $\modelname$ model and the $\blimgrec$ baseline. 
The instruction landmark accuracy converges to a relatively low value due to instruction ambiguity.

Our model enables us to easily visualize the agent perception and interpret the cause of errors. \figref{fig:fail_analysis} shows the visual process we can use to identify if the perception, grounding, planning, or control components failed.

\section{Discussion}
\label{sec:conclusion}

Neural network architectures have achieved remarkable performance in various high-level tasks, but their applications to the robotics domain have largely been limited to single, repeated tasks~\cite{levine2016end, giusti2016machine, mueller2017teaching} or required substantial amount of training data due to high sample-complexity~\cite{pinto2016supersizing}.

We show that a modular neural network architecture that (a) assigns explicit roles to its subcomponents in the form of auxiliary objectives; and (b) relieves the neural network from having to learn spatial transformations or memory operations that can be computed explicitly, can obtain strong performance on a complex visual navigation task that requires effective perception, symbol grounding, planning, and control.
The model is able to learn from limited amount of data, and generalize to unseen environments. 
Key to enabling this efficient learning is the combination of auxiliary objectives and a modular architecture that results in explicitly solving the symbol grounding problem~\cite{harnad1990symbol} by spatial reasoning on a high-level map representation.

There are several directions for future work that follow up on limitations of our model and setup. 
A key problem that is not addressed by our experiments is exploration. 
The wide field-of-view provided to the agent abstracts away issues of observability, and allows us to to focus on spatial reasoning and task-completion abilities. 
While the architecture is not specifically designed for fully observable environments, it is likely that the learning procedure will not be robust to such challenges. 
A second potential direction for future work is removing the auxiliary objectives. 
These objectives require ground truth labels of landmarks in the environment and meaning of the instruction. This type of information is challenging to obtain in physical environments or when using natural language instructions.

\section*{Acknowledgments}

We would like to thank Dipendra Misra, Ryan Benmalek, and Daniel Lee for helpful feedback and discussions. This research was supported by the Air Force Office of Scientific Research under award number FA9550-17-1-0109 and by Schmidt Sciences.  We are grateful for this support.

\bibliographystyle{plainnat}
\bibliography{references}

\end{document}